\documentclass[letterpaper, 10 pt, conference]{ieeeconf}  

\IEEEoverridecommandlockouts                              

\usepackage{graphics} 
\usepackage{epsfig} 
\usepackage{mathptmx} 
\usepackage{times} 
\usepackage{amsmath} 
\usepackage{amssymb}  
\usepackage{mathtools}
\usepackage{float}
\usepackage{tikz}
\usetikzlibrary{calc}

\title{\LARGE \bf
Prediction-Error-Tightened MPC for Socially Compliant Crowd Navigation
}

\author{
Alessandro Febbraro$^{1}$, Federico Rollo$^{2}$, Marcello Farina$^{1}$, Luca Bascetta$^{1}$ and Navvab Kashiri$^{2}$
\thanks{$^{1}$Politecnico di Milano, Dipartimento di Elettronica, Informazione e Bioingegneria, Milan, Italy {\tt\small {name.surname}@polimi.it}}%
\thanks{$^{2}$Robotics Laboratory, Innovation Hub \& Intellectual Property, Leonardo S.p.A., Genoa, Italy}
\thanks{\footnotesize
This work has been submitted to the IEEE for possible publication. Copyright may be transferred without notice, after which this version may no longer be accessible.
}
}

\begin{document}

\maketitle
\thispagestyle{empty}
\pagestyle{empty}

\begin{abstract}

Autonomous navigation of mobile robots in crowded environments, while preserving pedestrians’ psychological comfort, requires distinguishing pedestrians from other obstacles, predicting their future motion, and accounting for uncertainty in the predicted trajectories of both pedestrians and the robot. This work extends a previous MPC framework by integrating obstacle and pedestrian detection, Koopman-based motion prediction, and online prediction-error estimation directly within the collision-avoidance mechanism. To reduce collision-avoidance complexity in crowded scenarios, nearby obstacles are aggregated, thereby reducing the number of collision-avoidance constraints. Simulation studies evaluate the prediction-error margins and the computational benefits of semantics-aware aggregation. In a paired 300-seed closed-loop comparison with a moving generic obstacle, Markov residual tightening reduces the fraction of runs with physical contact from $10.0\%$ to $1.33\%$, at a mean travel-time cost of $0.68$~s. In a separate crowd-navigation semantic ablation, the offline robot-footprint/personal-space overlap occupies $2.11\%$ of observed time with pedestrian-aware handling versus $22.84\%$ with generic-obstacle handling.

\end{abstract}

\section{INTRODUCTION}

Mobile robots operating in crowded human environments must not only avoid collisions, but also move predictably and comfortably around pedestrians. This requires preserving suitable interpersonal distances, anticipating human motion, and avoiding abrupt maneuvers \cite{singamaneni2024survey,BOLDRER2022103979}, particularly for assistive platforms such as autonomous wheelchairs \cite{MORALES2017355}.

Socially aware navigation has been addressed through model-based, learning-based, and hybrid approaches \cite{singamaneni2024survey,Tolani2020VisualNA}. Learning-based methods can capture complex interactions but require substantial data, may generalize poorly to unseen scenarios, and typically lack explicit safety guarantees \cite{singamaneni2024survey,10811861,BOLDRER2022103979}. MPC offers an attractive model-based alternative because tracking, actuation limits, obstacle avoidance, motion smoothness, and proxemic requirements can be handled within a unified optimization framework \cite{SCHOELS20206555,9993397,11622475}. Its effectiveness, however, depends on reliable motion prediction and on maintaining real-time solvability as collision-avoidance constraints increase \cite{9993397}.

Prediction uncertainty has therefore received increasing attention in MPC-based navigation. Uncertainty-aware objectives and constraints can introduce conservatism depending on their formulation \cite{AKHTYAMOV2025104830}, while conformal-prediction methods construct uncertainty regions from offline calibration data \cite{lindemann2023safeplanningdynamicenvironments} or adapt them online from observed errors \cite{dixit2022adaptiveconformalpredictionmotion}. In this work, collision cones are used to identify relative velocities that may lead to a collision with an obstacle \cite{fiorini1993relativevelocity,tellarini2019autonomous,durante2018humanaware} and prediction errors are taken into account in both collision-cone detection and the collision-avoidance constraints of the MPC.

Short-horizon motion prediction remains challenging,
particularly when pedestrians change direction.
Koopman-based models approximate nonlinear motion
through linear evolution in a lifted space and have
been investigated for pedestrian and moving-obstacle
prediction \cite{ijcai2024p113,
bueno2025koopmanbaseddynamicenvironmentprediction}.
Here, a finite-dimensional Koopman operator is
updated online for each tracked dynamic obstacle.

A further challenge is computational scalability in crowds, where each obstacle may generate constraints over multiple prediction steps \cite{9993397}. Unlike group-level representations aimed primarily at modeling pedestrian-group motion \cite{wang2022groupbasedmotionpredictionnavigation}, we use grouping as a constraint-reduction mechanism. Nearby detections are aggregated while retaining pedestrian personal spaces and the detected geometry of generic obstacles; the resulting group representation is used both for collision-cone screening and for constructing a reduced set of separating constraints.

Using the linear-MPC/input--output-linearization architecture in
\cite{11622475} as the nominal control backbone, the main contributions
of this work are:
\begin{enumerate}
\item a prediction-error-tightened collision-avoidance formulation in which
errors affecting collision-cone screening and separating constraints are
characterized online through second moments and incorporated through
Markov-based margins; and
\item a semantics-aware obstacle-aggregation strategy that preserves
pedestrian personal spaces and generic-obstacle geometry while reducing
the number of separating collision-avoidance constraints.
\end{enumerate}
The first contribution is evaluated through margin-exceedance diagnostics and a paired closed-loop safety--efficiency study; the second is assessed via same-state QP benchmarks, multi-obstacle navigation runs, and an offline personal-space overlap assessment.

\section{NAVIGATION AND MOTION-PREDICTION FRAMEWORK}

\subsection{Robot Model and Control Architecture}

The architecture in~\cite{11622475} is adopted: an outer linear MPC
regulates the Cartesian position of a look-ahead point $P$, while
input--output linearization maps its virtual Cartesian velocity to the
longitudinal and angular velocities of the robot.
Denoting by $(x,y)$ the wheel-axis midpoint, by $\theta$ the robot
heading, and by $\epsilon>0$ the fixed look-ahead distance,
\[
    P=
    \begin{bmatrix}
        x+\epsilon\cos\theta &
        y+\epsilon\sin\theta
    \end{bmatrix}^{\top}.
\]
Defining
\[
    \boldsymbol{\xi}_k=
    \begin{bmatrix}
        x_{P,k} & y_{P,k}
    \end{bmatrix}^{\top},
    \qquad
    \mathbf{u}_k=
    \begin{bmatrix}
        v_{P_x,k} & v_{P_y,k}
    \end{bmatrix}^{\top},
\]
the exact discrete-time model of the linearized system under zero-order hold, with MPC
sampling period $T_s>0$, is
\begin{equation*}
    \boldsymbol{\xi}_{k+1}
    =
    \mathbf{A}\boldsymbol{\xi}_k
    +
    \mathbf{B}\mathbf{u}_k,
    \qquad
    \mathbf{A}=\mathbf{I}_2,
    \quad
    \mathbf{B}=T_s\mathbf{I}_2.
\end{equation*}
For a given heading $\theta$, the input--output-linearizing
transformation gives
\begin{equation*}
    v=
    \begin{bmatrix}
        \cos\theta & \sin\theta
    \end{bmatrix}\mathbf{u},
    \qquad
    \omega=
    \frac{1}{\epsilon}
    \begin{bmatrix}
        -\sin\theta & \cos\theta
    \end{bmatrix}\mathbf{u}.
\end{equation*}

\subsection{Nominal MPC Formulation}

At sampling instant $k$, the MPC optimizes
\begin{equation*}
    \mathbf{U}
    =
    \begin{bmatrix}
        \mathbf{u}_{k|k}^{\top} &
        \cdots &
        \mathbf{u}_{k+N-1|k}^{\top}
    \end{bmatrix}^{\top}
\end{equation*}
over a prediction horizon of length $N$ by minimizing
\begin{equation*}
    J_k
    =
    \sum_{i=0}^{N-1}
    \left(
        \left\|
            \boldsymbol{\xi}_{k+i+1|k}
            -
            \boldsymbol{\xi}_{k+i+1}^{\mathrm{ref}}
        \right\|_{\mathbf{Q}_i}^{2}
        +
        \left\|
            \mathbf{u}_{k+i|k}
        \right\|_{\mathbf{R}}^{2}
    \right),
\end{equation*}
where $\mathbf Q_i=\mathbf Q$ for
$i=0,\ldots,N-2$ and $\mathbf Q_{N-1}=\mathbf S$,
$\mathbf Q=q\mathbf I_2$ penalizes tracking error,
$\mathbf R=r\mathbf I_2$ penalizes virtual control, and
$\mathbf S=p\mathbf I_2$ is the terminal tracking weight
($q,r,p>0$).

The velocity-variation constraints in~\cite{11622475} are adopted,
whereas the former wheel-velocity limits are replaced by direct bounds
on the longitudinal and angular robot velocities:
\begin{equation}\label{eq:vel_const}
    -\bar v
    \leq
    \begin{bmatrix}
        \cos\bar{\theta}_{k+i|k-1} &
        \sin\bar{\theta}_{k+i|k-1}
    \end{bmatrix}
    \mathbf{u}_{k+i|k}
    \leq
    \bar v,
\end{equation}
\begin{equation}\label{eq:ang_vel_const}
    -\bar\omega
    \leq
    \frac{1}{\epsilon}
    \begin{bmatrix}
        -\sin\bar{\theta}_{k+i|k-1} &
        \cos\bar{\theta}_{k+i|k-1}
    \end{bmatrix}
    \mathbf{u}_{k+i|k}
    \leq
    \bar\omega,
\end{equation}
for $i=0,\ldots,N-1$. To preserve linearity,
$\bar{\theta}_{k|k-1}=\theta_k$, while $\bar{\theta}_{k+i|k-1}$ for $i\geq1$ is the heading
predicted at the previous MPC iteration.

\subsection{Track-Specific Koopman Motion Prediction}

For each tracked dynamic obstacle $j$, the observable vector is
\begin{equation*}
    \mathbf{z}_{j,k}
    =
    \begin{bmatrix}
        x_{j,k} &
        y_{j,k} &
        v_{x,j,k} &
        v_{y,j,k} &
        \theta_{j,k} &
        \omega_{j,k} &
        v_{j,k}
    \end{bmatrix}^{\top},
\end{equation*}
where
$v_{j,k}=\sqrt{v_{x,j,k}^{2}+v_{y,j,k}^{2}}$,
$\theta_{j,k}$ is the unwrapped direction of motion, and
$\omega_{j,k}$ its angular rate. Its evolution is approximated by
\begin{equation*}
    \mathbf{z}_{j,k+1}
    \simeq
    \mathbf{K}_{j,k}\mathbf{z}_{j,k},
\end{equation*}
where the track-specific operator is constrained to the following structure to reduce the number of free parameters while encoding simple kinematic relationships:
\begin{equation*}
\mathbf{K}_{j,k} =
\begin{bmatrix}
1 & k_{1,2} & k_{1,3} & k_{1,4} & k_{1,5} & k_{1,6} & k_{1,7} \\
k_{2,1} & 1 & k_{2,3} & k_{2,4} & k_{2,5} & k_{2,6} & k_{2,7} \\
0 & 0 & 1 & k_{3,4} & k_{3,5} & k_{3,6} & k_{3,7} \\
0 & 0 & k_{4,3} & 1 & k_{4,5} & k_{4,6} & k_{4,7} \\
0 & 0 & 0 & 0 & k_{5,5} & k_{5,6} & k_{5,7} \\
0 & 0 & 0 & 0 & 0 & k_{6,6} & k_{6,7} \\
0 & 0 & 0 & 0 & 0 & k_{7,6} & 1
\end{bmatrix}.
\end{equation*}
New tracks are initialized with a constant-translational-velocity and
constant-angular-rate prior, corresponding to an identity operator with
additional entries $T_s$ in positions $(1,3)$, $(2,4)$, and $(5,6)$.

Let $N_{\mathrm{s}}$ denote the number of stored samples. The operator
is identified online from a delayed window of $N_{\mathrm{tr}}$
one-step transitions,
\begin{equation*}
    \mathcal{W}_k
    =
    \{
        k-N_{\mathrm{s}}+1,
        \ldots,
        k-N_{\mathrm{s}}+N_{\mathrm{tr}}
    \}.
\end{equation*}

The oldest $N_{\mathrm{tr}}+1$ observations provide
$N_{\mathrm{tr}}$ one-step identification transitions.
Specifically,
\begin{equation*}
    \mathbf{K}_{j,k}
    =
    \operatorname*{arg\,min}_{\mathbf{K}\in\mathcal{K}}
    \sum_{\ell\in\mathcal{W}_k}
    \left\|
        \mathbf{z}_{j,\ell+1}
        -
        \mathbf{K}\mathbf{z}_{j,\ell}
    \right\|_2^2
    +
    \gamma
    \left\|
        \mathbf{K}
        -
        \mathbf{K}_{j,k-1}
    \right\|_F^2,
\end{equation*}
where $\mathcal{K}$ denotes the prescribed matrix structure and
$\gamma>0$ regularizes variations with respect to the previously
identified operator. The identified
operator is kept fixed over the current MPC horizon and propagated as
\begin{equation*}
    \hat{\mathbf{z}}_{j,k+h|k}
    =
    \mathbf{K}_{j,k}
    \hat{\mathbf{z}}_{j,k+h-1|k},
    \qquad
    h=1,\ldots,N,
\end{equation*}
with $\hat{\mathbf{z}}_{j,k|k}=\mathbf{z}_{j,k}$, $\hat{\mathbf{z}}_{j,k+h|k}$ is the prediction of the observable vector $\mathbf{z}_{j,k+h|k}$ at the prediction step $h$.
The predicted position and velocity are extracted from the first four
components.

\subsection{Nominal Collision-Threat Detection and Avoidance Constraints}

Let $G_m$ denote the set of member tracks in obstacle group $m$
(Section IV). At prediction step $h$, two tangent rays are drawn from the nominal predicted robot position to each member's semantic collision region. The two outermost rays define the group collision cone, used to identify potentially dangerous relative velocities \cite{fiorini1993relativevelocity,tellarini2019autonomous,durante2018humanaware}. This is a nominal collision-threat test, not a guarantee that a collision will occur. Denoting the selected tangent points by
$\mathbf{q}_{r,m,k+h|k}$, $r\in\{1,2\}$, and the nominal predicted
robot position by $\bar{\mathbf{p}}_{k+h|k}^{\mathrm R}$, define
\begin{equation*}
    \mathbf{d}_{r,m,k+h|k}
    =
    \frac{
        \mathbf{q}_{r,m,k+h|k}
        -
        \bar{\mathbf{p}}_{k+h|k}^{\mathrm{R}}
    }{
        \left\|
            \mathbf{q}_{r,m,k+h|k}
            -
            \bar{\mathbf{p}}_{k+h|k}^{\mathrm{R}}
        \right\|_2
    },
\end{equation*}
and
\begin{equation*}
    \mathbf{V}_{m,k+h|k}
    =
    \begin{bmatrix}
        \mathbf{d}_{1,m,k+h|k} &
        \mathbf{d}_{2,m,k+h|k}
    \end{bmatrix}.
\end{equation*}

Let $\bar{\mathbf v}_{k+h|k}^{\mathrm R}$ denote the predicted
planar robot velocity and $\hat{\mathbf v}_{j,k+h|k}$
the predicted velocity of obstacle $j$, both expressed
in the same reference frame. The mean relative velocity
between the robot and group $G_m$ is
\begin{equation*}
\bar{\mathbf v}_{m,k+h|k}^{\mathrm{rel}}
=
\frac{1}{|G_m|}
\sum_{j\in G_m}
\left(
\bar{\mathbf v}_{k+h|k}^{\mathrm R}
-
\hat{\mathbf v}_{j,k+h|k}
\right).
\end{equation*}

Following the collision-cone approach in
\cite{11622475}, this relative velocity is expressed
as a linear combination of the two unit vectors
defining the boundaries of the group collision cone:
\begin{equation*}
\bar{\mathbf v}_{m,k+h|k}^{\mathrm{rel}}
=
\mathbf V_{m,k+h|k}
\boldsymbol{\alpha}_{m,k+h|k},
\end{equation*}
where
$\boldsymbol{\alpha}_{m,k+h|k}
=
[\alpha_{1,m,k+h|k},\alpha_{2,m,k+h|k}]^\top$.

If the two tangent directions are linearly independent,
the coefficients are uniquely determined. If both
coefficients are nonnegative, the relative velocity lies
inside the collision cone, indicating a potential
collision:
\begin{equation*}
\alpha_{1,m,k+h|k}\geq 0,
\qquad
\alpha_{2,m,k+h|k}\geq 0.
\end{equation*}

A potential collision threat is also detected when the
predicted robot position lies inside the aggregate
collision region, independently of the coefficient test. The latter case is checked separately
because the ray decomposition is not defined inside the region.

For a retained group, Section IV constructs a separating line with
normal $\mathbf{h}_{m,k+h|k}$ and scalar offset
$\ell_{m,k+h|k}$. The nominal
collision-avoidance constraint is
\begin{equation*}
    \mathbf{h}_{m,k+h|k}^{\top}
    \boldsymbol{\xi}_{k+h|k}
    \leq
    \ell_{m,k+h|k},
\end{equation*}
which is linear in the MPC decision variables through the prediction
model. Feasibility is handled through
\begin{equation*}
\begin{aligned}
    \mathbf{h}_{m,k+h|k}^{\top}\boldsymbol{\xi}_{k+h|k}
    &\leq
    \ell_{m,k+h|k}
    +d_{m,k+h|k}^{\min}s_{m,k},\\
    0&\leq s_{m,k}\leq1,
\end{aligned}
\end{equation*}
where $d_{m,k+h|k}^{\min}$ is the geometry-dependent maximum
relaxation and $s_{m,k}$ is a bounded slack variable. 
Following the approach in \cite{11622475},
the slack variables $s_{m,k}$, together with those
of the velocity-variation constraints, are
quadratically penalized in the MPC cost function
to discourage constraint violations:
\begin{equation*}
\widetilde J_k
=
J_k
+
\|\mathbf s_k\|_{\mathbf S_s}^{2}
+
\|\mathbf s_{a,k}\|_{\mathbf S_a}^{2},
\end{equation*}
where $\mathbf S_s$ and $\mathbf S_a$ are the
corresponding penalty weight matrices.
Section III introduces prediction-error-dependent
modifications to both the collision-cone screening and the separating
constraint.
\section{UNCERTAINTY-AWARE COLLISION AVOIDANCE}

\subsection{Online Characterization of Multi-Step Prediction Errors}

The uncertainty margins are estimated online by historical backtesting
directly at the scalar quantities used by the collision-avoidance
mechanism, rather than by separately modeling robot and obstacle state
uncertainty. Hence, the resulting errors jointly account for prediction
inaccuracies and their effect on the collision geometry.

At sampling instant $k$, let $\mathcal{J}_m(k)$ be the set of indexes of the tracks currently assigned to $G_m$, and let
$N_{\mathrm e}$
denote the number of historical samples retained for uncertainty
estimation. Since groups may split or
merge, this membership is kept fixed during backtesting instead of
treating the group index as a persistent identity. In the adopted implementation, the stored
observation history is partitioned into
identification and backtesting portions:
\begin{equation*}
    N_{\mathrm{s}}
    =
    \underbrace{(N_{\mathrm{tr}}+1)}_{\text{Koopman}}
    +
    \underbrace{N}_{\text{look-back}}
    +
    \underbrace{N_{\mathrm e}}_{\text{targets}}.
\end{equation*}
The oldest $N_{\mathrm{tr}}+1$ states are used
for Koopman identification, as described
in Section~II-C. The next $N$
states provide the additional history needed
for the longest backtesting horizon, while
the most recent $N_{\mathrm e}$ states are
the comparison targets. For each prediction
step $h\in\{1,\ldots,N\}$ and historical sample
$q\in\{0,\ldots,N_{\mathrm e}-1\}$, an $h$-step prediction is
reconstructed from origin $k-q-h$ to target $k-q$. The current
track-specific Koopman model is propagated from the stored obstacle
state, while the robot trajectory is reconstructed from the stored
robot state and MPC control sequence at the same prediction origin.
Predicted and observed collision-related quantities are then recomputed
through the same group-level construction using $\mathcal{J}_m(k)$.

Let $\psi$ denote a scalar collision-related quantity: either a
collision-cone coefficient $\alpha_r$ or a separating-constraint
residual $g$. For group $m$ and horizon $h$, its historical prediction
error is the observed value minus the past prediction:
\begin{equation}\label{eq:generic_prediction_error}
    \Delta\psi_{m,h}^{(q)}
    =
    \psi_{m,k-q}^{\mathrm{obs}}
    -
    \hat{\psi}_{m,k-q|k-q-h},
\end{equation}
and its empirical raw second moment is
\begin{equation}\label{eq:empirical_second_moment}
    \widehat{M}_{2,\psi,m,h}(k)
    =
    \frac{1}{N_{\mathrm e}}
    \sum_{q=0}^{N_{\mathrm e}-1}
    \left(
        \Delta\psi_{m,h}^{(q)}
    \right)^2.
\end{equation}

For a generic error $\Delta\psi$ with finite true second moment
$M_{2,\psi}=\mathbb{E}[(\Delta\psi)^2]$ and threshold $d_\psi>0$,
Markov's inequality applied to $(\Delta\psi)^2$ gives
\begin{equation}\label{eq:markov_generic}
    \mathbb{P}\!\left(
        |\Delta\psi|\geq d_\psi
    \right)
    \leq
    \frac{M_{2,\psi}}{d_\psi^2}.
\end{equation}
Choosing $d_\psi$ so that the upper bound in
\eqref{eq:markov_generic} equals
$\varepsilon_\psi$ yields
\begin{equation}\label{eq:markov_margin_generic_theoric}
    d_\psi
    =
    \sqrt{
        \frac{M_{2,\psi}}{\varepsilon_\psi}
    },
    \qquad
    \varepsilon_\psi\in(0,1).
\end{equation}
Replacing the unknown true second moment with
its online estimate $\widehat{M}_{2,\psi}$
yields the empirical margin
\begin{equation}\label{eq:markov_margin_generic}
    d_\psi
    =
    \sqrt{
        \frac{\widehat{M}_{2,\psi}}{\varepsilon_\psi}
    },
    \qquad
    \varepsilon_\psi\in(0,1).
\end{equation}
Replacing an expectation by a finite-window estimate does not,
by itself, provide a finite-sample probabilistic guarantee. Grouping uncertainty
is not modeled: the characterization is conditioned on the group
composition available at the current sampling instant.

\subsection{Uncertainty-Aware Collision-Cone Screening}

The construction above is first applied to the two collision-cone
coefficients. For $r\in\{1,2\}$,
\begin{equation}\label{eq:alpha_prediction_error}
    \Delta\alpha_{r,m,h}^{(q)}
    =
    \alpha_{r,m,k-q}^{\mathrm{obs}}
    -
    \hat{\alpha}_{r,m,k-q|k-q-h}.
\end{equation}
Here $\alpha^{\mathrm{obs}}$ is reconstructed from measured states
and $\hat\alpha$ from the corresponding earlier prediction. Since
the cone decomposition is defined only outside the aggregate
collision region, let $\mathcal{Q}_{\alpha,m,h}(k)$ contain the
historical samples for which both predicted and observed robot
positions lie outside their corresponding aggregate regions. For
$|\mathcal{Q}_{\alpha,m,h}(k)|>0$,
\begin{equation}\label{eq:alpha_second_moment}
\begin{split}
    \widehat{M}_{2,\alpha_r,m,h}(k)
    &=
    \frac{1}{|\mathcal{Q}_{\alpha,m,h}(k)|}
    \sum_{q\in\mathcal{Q}_{\alpha,m,h}(k)}
    \left(
        \Delta\alpha_{r,m,h}^{(q)}
    \right)^2,\\
    d_{\alpha_r,m,h}(k)
    &=
    \sqrt{
        \frac{
            \widehat{M}_{2,\alpha_r,m,h}(k)
        }{
            \varepsilon_\alpha
        }
    },
    \qquad
    r\in\{1,2\},
\end{split}
\end{equation}
where $\varepsilon_\alpha\in(0,1)$.

Since $|\Delta\alpha_r|\leq d_{\alpha_r}$ implies
$\alpha_r^{\mathrm{obs}}\leq\hat{\alpha}_r+d_{\alpha_r}$, a group is
retained as a potential collision threat whenever
\begin{equation}\label{eq:uncertain_collision_cone_test}
\begin{aligned}
    \alpha_{1,m,k+h|k}
    +d_{\alpha_1,m,h}(k)
    &\geq0,\\
    \alpha_{2,m,k+h|k}
    +d_{\alpha_2,m,h}(k)
    &\geq0.
\end{aligned}
\end{equation}
A group failing either corrected coefficient is screened out,
subject to the speed and inside-region tests of Section~II-D.
When no eligible historical errors exist, the implementation sets
$d_{\alpha_r}=0$: no error-based correction is available for that
case. Larger historical errors generally increase the margins and
make screening more conservative.

\subsection{Uncertainty-Tightened Separating Constraints}

For a retained group, prediction errors affect both the robot position
and the separating-line coefficients. Their combined effect is
characterized through the constraint residual. For the $q$-th
historical $h$-step prediction, define
\begin{equation}\label{eq:g_residuals}
\begin{aligned}
    \hat g_{m,k-q|k-q-h}
    &=
    \hat\ell_{m,k-q|k-q-h}
    -
    \hat{\mathbf h}_{m,k-q|k-q-h}^{\top}
    \bar{\boldsymbol\xi}_{k-q|k-q-h},\\
    g_{m,k-q}^{\mathrm{obs}}
    &=
    \ell_{m,k-q}^{\mathrm{obs}}
    -
    \left(
        \mathbf h_{m,k-q}^{\mathrm{obs}}
    \right)^{\top}
    \boldsymbol\xi_{k-q}^{\mathrm{obs}},
\end{aligned}
\end{equation}
where $\bar{\boldsymbol\xi}$ denotes the nominal predicted
look-ahead-point position, the superscript $\mathrm{obs}$ denotes
quantities reconstructed at the observed state, and hats mark
quantities from the earlier prediction. The separating geometry is
recomputed for both states using the same tracks. The residual
prediction error is
\begin{equation}\label{eq:g_prediction_error}
    \Delta g_{m,h}^{(q)}
    =
    g_{m,k-q}^{\mathrm{obs}}
    -
    \hat g_{m,k-q|k-q-h}.
\end{equation}
Accordingly,
\begin{equation}\label{eq:g_margin}
\begin{split}
    \widehat{M}_{2,g,m,h}(k)
    &=
    \frac{1}{N_{\mathrm e}}
    \sum_{q=0}^{N_{\mathrm e}-1}
    \left(
        \Delta g_{m,h}^{(q)}
    \right)^2,\\
    d_{g,m,h}(k)
    &=
    \sqrt{
        \frac{
            \widehat{M}_{2,g,m,h}(k)
        }{
            \varepsilon_g
        }
    },
    \qquad
    \varepsilon_g\in(0,1).
\end{split}
\end{equation}

Because $g^{\mathrm{obs}}=\hat g+\Delta g$, imposing
$\hat g\geq d_g$ preserves a nonnegative residual for errors satisfying
$|\Delta g|\leq d_g$. The corresponding tightened constraint is
\begin{equation}\label{eq:tightened_constraint}
    \mathbf h_{m,k+h|k}^{\top}
    \boldsymbol\xi_{k+h|k}
    \leq
    \ell_{m,k+h|k}
    -
    d_{g,m,h}(k).
\end{equation}
Combining this tightening with the geometry-dependent relaxation of
Section II-D gives
\begin{equation}\label{eq:tightened_soft_constraint}
\begin{split}
    \mathbf h_{m,k+h|k}^{\top}
    \boldsymbol\xi_{k+h|k}
    \leq{}&
    \ell_{m,k+h|k}
    -
    d_{g,m,h}(k)
    \left(
        1-s_{m,k}
    \right)
    \\
    &+
    d^{\min}_{m,k+h|k}s_{m,k},
    \qquad
    0\leq s_{m,k}\leq1.
\end{split}
\end{equation}
Hence, $s_{m,k}=0$ yields the fully tightened constraint, whereas
$s_{m,k}=1$ recovers the geometry-dependent minimum-separation
condition. The two uncertainty margins act at successive stages:
$d_{\alpha_r,m,h}(k)$ modifies collision-threat screening and
$d_{g,m,h}(k)$ tightens the constraint subsequently imposed by the MPC.
\section{PERCEPTION AND SEMANTICS-AWARE OBSTACLE AGGREGATION}

\subsection{Semantic Perception and Obstacle Representation}

LiDAR detections are tracked as circles or line segments and converted
into polygonal collision regions before grouping. A detected line segment is represented by a rectangle obtained by inflating the detected segment by $3\,\mathrm{cm}$, whereas a detected circle is replaced by a circumscribed regular decagon. For a generic
obstacle $i$, let $\{\mathbf w_{i,s}\}$ denote the vertices of the resulting
polygon. This polygonal description is retained throughout grouping and is
used directly in the support and separating-line computations of
Section~IV-C. No pedestrian-specific proxemic enlargement is assigned to
generic obstacles; their clearance from the robot is accounted for later
through the directional robot-footprint margin.

\begin{figure}[h!]
    \centering
    \includegraphics[width=0.85\columnwidth]{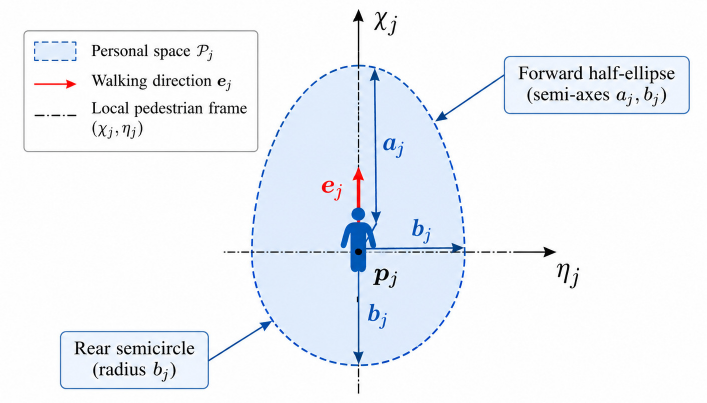}
    \caption{Asymmetric pedestrian personal space.}
\label{fig:personal_space}
\end{figure}

The nominal perception pipeline identifies pedestrians through
camera--LiDAR fusion using a YOLO-family image detector
\cite{redmon2016yolo} and projected LiDAR points; the resulting detections are
associated with the LiDAR tracks in the map frame. Their collision
representation is then replaced by an asymmetric, velocity-dependent
personal space of the type used in \cite{11622475}. Consider a
pedestrian-centered local frame with longitudinal coordinate $\chi_j$
aligned with the walking direction $\mathbf e_j$ and pointing forward,
and lateral coordinate $\eta_j$. The personal space is the union of a rear
semicircular region and a forward half-ellipse,
\begin{equation}\label{eq:personal_space_geometry}
\mathcal P_j=
\left\{
(\chi_j,\eta_j)\in\mathbb R^2 \;\middle|\;
\begin{array}{ll}
\displaystyle
\frac{\chi_j^2}{b_j^2}+\frac{\eta_j^2}{b_j^2}\leq1,
& \chi_j<0,\\[2mm]
\displaystyle
\frac{\chi_j^2}{a_j^2}+\frac{\eta_j^2}{b_j^2}\leq1,
& \chi_j\geq0
\end{array}
\right\}.
\end{equation}
The forward semi-axis is set to $a_j=1.5b_j$,
where $b_j$ is the lateral semi-axis and rear
semicircle radius. Denoting by
$\mathbf{v}_{\mathrm{rel},j}
=\mathbf{v}_R-\mathbf{v}_j$
the robot--pedestrian relative velocity, the
lateral semi-axis is defined as
\begin{equation}
b_j=\min\left\{1.5,\max\left(
r_c,3\|\mathbf{v}_{\mathrm{rel},j}\|_2
\right)\right\},
\end{equation}
where $r_c$ is the minimum personal-space radius.
In the absolute reference frame, the region is rotated by the pedestrian heading $\theta_j$ and translated to its predicted position
$\mathbf{p}_j$ (Fig.~\ref{fig:personal_space}). This geometry is preserved
during grouping for collision-cone and separating-
constraint construction (Section~IV-C).

\subsection{Semantics-Aware Obstacle Grouping}

To reduce the number of MPC collision constraints, detections are first
grouped through an undirected proximity graph. Two circle/segment
primitives are connected whenever their minimum geometric clearance is
below a threshold $\tau_{\mathrm g}$, and each connected component defines
an initial group $G_m$. The initial connected components have no
fixed size cap; when a group must be subdivided, the splitting
routine limits mergers by group size. Segments exceeding the
implementation's long-segment threshold are retained as individual
groups to prevent extended structures from merging otherwise
distinct obstacles. The
semantic representation of each member is preserved when evaluating the
aggregate collision geometry. If an aggregate region contains either the current controlled point or the navigation reference, the grouping routine attempts a finer
subdivision and repeats the consistency check whenever a nontrivial
split is available.

\subsection{Group-Level Collision Representation}

For each group $G_m$, the inside-region test is constructed from
$N_{\mathrm a}$ uniformly distributed directions
$\mathbf n_\nu=[\cos\varphi_\nu\ \sin\varphi_\nu]^\top$, with $\nu=0,\ldots,N_{\mathrm a}-1$ and
$\varphi_\nu=2\pi\nu/N_{\mathrm a}$. Here $\nu$ labels sampled directions, while $\ell$ will denote
a separating-line offset. For a pedestrian $j$, using the personal-space geometry in \eqref{eq:personal_space_geometry} and Fig.~\ref{fig:personal_space}, let
$\mathbf p_j$ and $\mathbf e_j$ denote its predicted position and heading
direction and define
$\mathbf B_j=\mathcal{R}(\theta_j)\operatorname{diag}(a_j,b_j)$,
where $\mathcal{R}(\theta)$ denotes a planar rotation and
$\operatorname{diag}(a_j,b_j)$ the diagonal matrix with entries
$a_j,b_j$. The support (maximum projection along a direction) is
\begin{equation}
 s_j(\mathbf n_\nu)=\mathbf n_\nu^\top\mathbf p_j+
 \begin{cases}
   \|\mathbf B_j^\top\mathbf n_\nu\|_2,
   & \mathbf n_\nu^\top\mathbf e_j\geq0,\\
   b_j, & \mathbf n_\nu^\top\mathbf e_j<0,
 \end{cases}
\end{equation}
whereas generic-obstacle supports are computed directly from the
polygonal collision regions described in Section~IV-A. Let $\bar{\mathbf p}^{\mathrm R}$ denote the nominal predicted
robot position in the same frame. Omitting prediction
indices for compactness, the sampled group envelope and inside test
are
\begin{equation}\label{eq:group_support}
 \begin{aligned}
 s_m(\mathbf n_\nu)&=\max_{j\in G_m}s_j(\mathbf n_\nu),\\
 \mathbf n_\nu^\top\bar{\mathbf p}^{\mathrm R}
 &\leq s_m(\mathbf n_\nu),\qquad \forall\nu.
 \end{aligned}
\end{equation}
This sampled support representation is used only for the inside-region
test. The collision-cone boundaries are constructed separately by
computing two tangent rays for each semantic collision region and
retaining the two angularly extreme rays over $G_m$; their tangent points
are the $\mathbf q_{1,m,k+h|k}$ and $\mathbf q_{2,m,k+h|k}$ used in
Section II-D.
Thus, the group-level cone is constructed from the individual semantic collision regions rather than from a single surrogate group obstacle.

For a retained group, a single separating line is constructed. Omitting
group and prediction-step indices, let
\begin{equation}\label{eq:separator_initial}
 \begin{aligned}
 \mathbf c&=\frac{1}{|G_m|}\sum_{j\in G_m}\mathbf p_j,
 &\mathbf h^0&=\frac{\mathbf c-\bar{\mathbf p}^{\mathrm R}}
 {\|\mathbf c-\bar{\mathbf p}^{\mathrm R}\|_2},\\
 \mathbf h&=\mathcal{R}\!\left(\frac{\pi}{6}\delta_{\mathrm{dev}}\right)\mathbf h^0,
 &\ell^0&=\mathbf h^\top\mathbf c,
 \end{aligned}
\end{equation}
where $\delta_{\mathrm{dev}}\in\{-1,1\}$ identifies the selected bypass
side. The directional robot-footprint margin is
\begin{equation}\label{eq:robot_directional_margin}
 \delta^{\mathrm R}(\mathbf h)=
 \max\!\left\{\delta_0,
 \max_s\mathbf h^\top(\mathbf w_s^{\mathrm R}-\bar{\boldsymbol\xi})
 \right\},
\end{equation}
where $\mathbf w_s^{\mathrm R}$ are the predicted robot-footprint
vertices and $\delta_0>0$ is a minimum geometric margin.
Writing $\delta^{\mathrm R}=\delta^{\mathrm R}(\mathbf h)$ and defining
$\widetilde{\mathbf B}_i=\mathcal{R}(\theta_i)
\operatorname{diag}(a_i+\delta^{\mathrm R},b_i+\delta^{\mathrm R})$,
let $\mathbf w_{i,s}$ denote the $s$-th vertex of the polygonal
collision region of generic obstacle $i$. The member margin is
\begin{equation}\label{eq:member_margin}
\begin{aligned}
 m_i={}&
 \begin{cases}
 \mathbf h^\top\mathbf p_i-\ell^0-b_i-\delta^{\mathrm R},
 & i\ \mathrm{ped.},\ \mathbf h^\top\mathbf e_i\geq0,\\
 \mathbf h^\top\mathbf p_i-\ell^0-
 \|\widetilde{\mathbf B}_i^\top\mathbf h\|_2,
 & i\ \mathrm{ped.},\ \mathbf h^\top\mathbf e_i<0,\\
 \begin{aligned}
 \mathbf h^\top\mathbf p_i-\ell^0-\delta^{\mathrm R}\\[-1mm]
 {}+\min_s\mathbf h^\top(\mathbf w_{i,s}-\mathbf p_i)
 \end{aligned}
 & i\ \mathrm{generic}.
 \end{cases}
\end{aligned}
\end{equation}
The pedestrian branch is reversed with respect to the support
construction because $m_i$ depends on the minimum projection along
$\mathbf h$, equivalently on the support in direction $-\mathbf h$.

Let $i^\star=\arg\min_{i\in G_m}m_i$ and
$m^\star=m_{i^\star}$. The final offset is
\begin{equation}\label{eq:line_translation}
 \ell=\ell^0+\min\{0,m^\star\},
\end{equation}
so the line is translated only when the most critical collision region
intersects the initial half-space. Let $r_{\mathrm{red}}>0$ denote the
reduced separation distance used under maximum relaxation. The
geometry-dependent relaxation is
\begin{equation}\label{eq:dmin_computation}
 d^{\min}=\begin{cases}
 \mathbf h^\top\mathbf p_{i^\star}-\ell-r_{\mathrm{red}},
 & m^\star\geq0\ \text{or}\ i^\star\ \mathrm{ped.},\\
 \delta^{\mathrm R}(\mathbf h),
 & m^\star<0\ \text{and}\ i^\star\ \mathrm{generic}.
 \end{cases}
\end{equation}
The centroid-based separator is accepted if
\begin{equation}\label{eq:dist_temp}
 d_{\mathrm{sep}}
 =\mathbf h^\top\bar{\boldsymbol\xi}-\ell-d^{\min}\leq0.
\end{equation}

Otherwise, a discrete search is performed over
$\mathbf h_\nu=-\mathbf n_\nu$. For each candidate, the line offset,
robot-footprint margin, member margins, and $d_\nu^{\min}$ are recomputed as
above. Isolated long segments are evaluated directly using the unrelaxed
separating-line residual
$\mathbf h_\nu^\top\bar{\boldsymbol\xi}-\ell_\nu$, whereas the centroid
fallback uses the maximally relaxed residual
$\mathbf h_\nu^\top\bar{\boldsymbol\xi}-\ell_\nu-d_\nu^{\min}$. The
minimum-residual direction is selected; for non-long-segment groups this
search is activated only when \eqref{eq:dist_temp} fails. The corresponding
$\mathbf h_{m,k+h|k}$, $\ell_{m,k+h|k}$, and
$d^{\min}_{m,k+h|k}$ are then used by the MPC.

\section{SIMULATION EVALUATION}
\subsection{Simulation Setup and Evaluation Metrics}

Three components assess prediction-error margins,
paired closed-loop safety and efficiency, and the benefits of
semantics-aware aggregation. In the single-obstacle ROS~2/Gazebo studies, a skid-steering
robot passes a moving cube classified either as a pedestrian or a
generic obstacle: only its semantic collision geometry changes.
The cube's nominal velocity is $(-0.30,0.00)\,\mathrm{m/s}$;
Gauss--Markov acceleration disturbances have standard deviation
$0.10\,\mathrm{m/s^2}$ per axis, correlation $\rho=0.90$, and
bound $0.30\,\mathrm{m/s^2}$ per axis. We use $T_s=0.2\,\mathrm{s}$,
$N=10$, $a_{\max}=0.2\,\mathrm{m/s^2}$, $N_{\mathrm{tr}}=20$,
$N_{\mathrm{s}}=51$, and $N_{\mathrm e}=20$.

For residual tightening, $d_g=0$ (baseline) is compared with
$\varepsilon_g\in\{0.05,0.10,0.20,0.30\}$ at fixed
$\varepsilon_\alpha=0.5$; semantic geometry and cone screening
remain active in the baseline. Each of $300$ attempted runs per setting has a deterministic
warm-up followed by a scheduled stochastic onset at $T_0$.
We evaluate $j=0,\ldots,90$ and $h=1,\ldots,8$
($0.2$--$1.6\,\mathrm{s}$), excluding runs with incomplete or
unsynchronized windows from \emph{shadow} statistics only.
A separate, paired 300-seed study compares active generic-obstacle
controllers with $\varepsilon_g=0.05$ and $d_g=0$, preserving the
same prescribed obstacle disturbances and other controller settings.
Navigation is evaluated using Nav2 goal status, ground-truth path
length, time to goal, minimum planar rectangle-based clearance,
QP statuses, and slack activity. Physical Husky--cube contacts are counted from
full-run Gazebo contact-sensor logs, independently
of the shadow-window validity checks.
In implemented cone screening, coefficient-based threat detection
requires relative speed above $0.1\,\mathrm{m/s}$; the separate
inside-region test does not impose this threshold.

\subsection{Empirical Evaluation of Prediction-Error Margins}

Residual tightening and collision-cone screening are assessed separately.

\subsubsection{Separating-Constraint Residual Margins}

For semantic class $s\in\{\mathrm{gen},\mathrm{ped}\}$, setting
$c\in\{\mathrm{baseline},0.05,0.10,0.20,0.30\}$, valid run $r$,
origin $j$, and horizon $h$, define the one-sided exceedance indicator
\begin{equation}
\label{eq:directional_exceedance}
I_{r,j,h}^{s,c}
=
\mathbf{1}
\left\{
\Delta g_{r,j,h}^{s,c}
<
-d_{g,r,j,h}^{s,c}
\right\},
\end{equation}
and average it over $R_{s,c}$ valid runs:
\begin{equation}
\label{eq:empirical_vr}
\widehat{\mathrm{VR}}_{h}^{s,c}(j)
=
\frac{1}{R_{s,c}}
\sum_{r=1}^{R_{s,c}}
I_{r,j,h}^{s,c}.
\end{equation}
In the baseline $d_g=0$, so exceedance means $\Delta g<0$;
otherwise the online margin follows \eqref{eq:g_margin}.
Thus, VR measures margin exceedances, not physical collisions.

In the independent pedestrian/generic campaigns,
baseline VR uses $293$ pedestrian and $284$ generic
valid runs, versus $299$ and $287$ at
$\varepsilon_g=0.05$, respectively;
$100\varepsilon_g\%$ is a design level,
not a probabilistic guarantee.
At $j=0$, both baselines reach $100\%$ VR for every horizon;
subsequent rates vary with time. Different semantic geometries
prevent interpreting their VR differences as a safety ranking.

A separate generic-obstacle \emph{shadow} study
used the active Markov controller at
$\varepsilon_g=0.05$ ($300$ attempts, $298$ valid).
Fifty seed-indexed runs calibrated the fixed-margin
comparator by horizon and design level.
The four adaptive methods (Markov, Cantelli,
Gaussian, empirical quantile) instead used the
same $N_{\mathrm e}=20$ rolling historical residuals.
All methods were evaluated on $248$ shared held-out
runs ($180{,}544$ origin--horizon pairs) at
$\varepsilon_g\in\{0.05,0.10,0.20,0.30\}$,
using trajectories generated by the active
Markov controller.

\begin{figure*}[!t]
    \centering
    \includegraphics[width=0.75\textwidth]
    {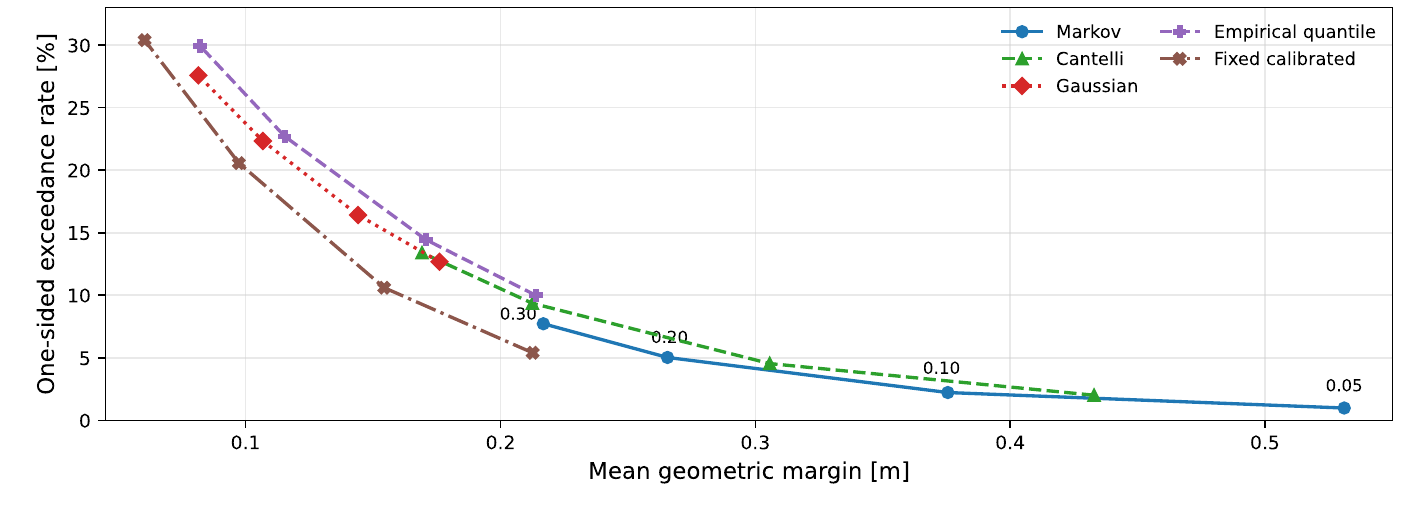}
    \caption{Paired shadow: one-sided residual
    exceedance rate versus mean geometric margin;
    comparators evaluated offline on $248$ shared runs.}
    \label{fig:vr_shadow_tradeoff}
\end{figure*}

Table~\ref{tab:vr_runwise_summary} summarizes run-level \emph{VR}
from earlier independent controller configurations; unlike the
paired shadow comparison, their trajectories and valid-run sets differ.
These VR values are not physical-contact frequencies.

\begin{table}[!ht]
    \centering
    \caption{Run-level VR means, variances, and valid runs.}
    \label{tab:vr_runwise_summary}
    \scriptsize
    \setlength{\tabcolsep}{2.0pt}
    \begin{tabular}{lrrrrrr}
    \hline
    & \multicolumn{2}{c}{Mean VR [\%]}
    & \multicolumn{2}{c}{Variance [\%$^2$]}
    & \multicolumn{2}{c}{Valid runs} \\
    Configuration & Ped. & Generic & Ped. & Generic & Ped. & Generic \\
    \hline
    $d_g=0$ & 52.33 & 52.21 & 110.41 & 135.95 & 293 & 284 \\
    $\varepsilon_g=0.05$ & 1.04 & 0.90 & 0.73 & 0.70 & 299 & 287 \\
    $\varepsilon_g=0.10$ & 2.39 & 2.37 & 2.63 & 3.04 & 297 & 289 \\
    $\varepsilon_g=0.20$ & 4.99 & 5.45 & 7.06 & 11.00 & 299 & 282 \\
    $\varepsilon_g=0.30$ & 7.38 & 8.18 & 11.51 & 17.45 & 299 & 286 \\
    \hline
    \end{tabular}
\end{table}

Fig.~\ref{fig:vr_shadow_tradeoff} compares Markov
with Cantelli (sample variance), Gaussian,
rolling empirical-quantile, and horizon-calibrated
fixed margins. At $\varepsilon_g=0.05$,
exceedance rates (mean margins) are:
Markov $0.99\%$ ($0.531$~m),
Cantelli $2.01\%$ ($0.433$~m),
Gaussian $12.68\%$ ($0.176$~m),
empirical quantile $9.99\%$ ($0.214$~m),
and fixed $5.40\%$ ($0.213$~m).
The Markov run-bootstrap $95\%$ interval is
$[0.87,1.12]\%$.
Its margin is at least the largest historical
adverse error for $N_{\mathrm e}=20$ and
$\varepsilon_g=0.05$.
At stochastic onset ($j=0$), exceedance exceeds
$99\%$ at $h=7,8$; all aggregates retain
these values. These offline results do not
establish closed-loop benefits for comparator
controllers or finite-sample coverage guarantees.

\subsubsection{Collision-Cone Screening Margins}

For each class $s\in\{\mathrm{gen},\mathrm{ped}\}$, zero cone
margins are compared with $\varepsilon_\alpha=0.5$, keeping
residual tightening disabled ($d_g=0$). The numbers of valid runs are $288$ (zero)
and $290$ (Markov) for generic obstacles,
and $295$ in both pedestrian configurations,
out of $300$ attempts each; $\mathcal V_{s,c}$ denotes the independently selected valid
runs. Let $D_{r,j,h}^{s,c}=1$ denote a group retained by screening and
$C_{r,j,h}^{s,c}=1$ a potential threat reconstructed from
subsequent MPC observations. For each horizon, MCAP pools
the $91$ prediction origins:
\begin{equation}\label{eq:mcap_cone}
\widehat{\mathrm{MCAP}}_h^{s,c}
=\frac{\displaystyle\sum_{r\in\mathcal V_{s,c}}\sum_{j=0}^{90}
\mathbf1\{D_{r,j,h}^{s,c}=0,\,C_{r,j,h}^{s,c}=1\}}
{\displaystyle\sum_{r\in\mathcal V_{s,c}}\sum_{j=0}^{90}
\mathbf1\{D_{r,j,h}^{s,c}=0\}}.
\end{equation}
The ratio is undefined if there are no screened-out cases.
Observed threats include the speed gate of Section~II-D and
do not represent physical collisions.
\begin{figure}[!t]
    \centering
    \includegraphics[width=0.85\columnwidth]
    {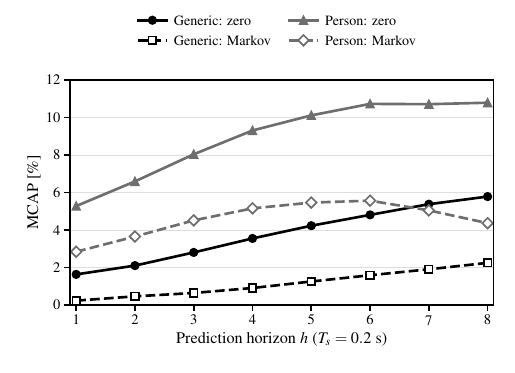}
    \caption{Collision-cone MCAP versus horizon
    for generic obstacles and pedestrians.}

    \label{fig:mcap_by_h}
\end{figure}
Fig.~\ref{fig:mcap_by_h} shows the horizon-wise MCAP.
Pooling all $(j,h)$, MCAP falls from $3.81\%$ to $1.11\%$
(generic) and from $9.00\%$ to $4.54\%$ (person).
Run-level bootstrap $95\%$ intervals for the reductions are
$[1.98,3.42]$ and $[3.47,5.45]$ percentage points,
respectively. The screened-out fractions fall from
$70.36\%$ to $57.69\%$ and from $59.97\%$ to $43.98\%$,
while false-activation fractions ($D=1,C=0$, over all valid
pairs) rise from $5.89\%$ to $16.54\%$ and from $5.51\%$
to $19.37\%$, respectively.
Thus, fewer observed threats are missed at the cost of more
conservative screening.

\subsection{Paired Closed-Loop Safety and Efficiency}
\label{sec:closedloop_safety}

We compare $300$ matched-seed generic-obstacle navigations
with active $\varepsilon_g=0.05$ or $d_g=0$, using identical
prescribed cube disturbances. Nav2-status and contact-log checks pass
for all $600$ trials. A contact run contains at least one
Gazebo-confirmed Husky--cube contact, independently of goal
success and of shadow residual exceedances.

\begin{table}[!htbp]
\centering
\caption{Paired closed-loop results (300 seeds); clearance
is a planar-geometry proxy. Means are across runs.}
\label{tab:paired_closedloop_300}
\scriptsize
\setlength{\tabcolsep}{3.0pt}
\begin{tabular}{lrr}
\hline
Metric & Markov & $d_g=0$ \\
\hline
Contact runs & $4/300$ & $30/300$ \\
Nav2 goal success & $300/300$ & $300/300$ \\
Time to goal [s], mean & 23.00 & 22.32 \\
Path length [m], mean & 11.296 & 11.006 \\
Minimum planar clearance [m], mean & 1.238 & 0.989 \\
Failed QPs / post-$T_0$ solves & $19/35{,}091$ & $187/34{,}032$ \\
Slack-active QPs [\%], run mean & 7.49 & 5.63 \\
Freezing $\geq5$ s / evaluable & $0/300$ & $1/293$ \\
\hline
\end{tabular}
\end{table}

Contact-run incidence decreases from $10.00\%$
to $1.33\%$: $3$ seeds contact in both modes,
$1$ only with Markov, and $27$ only with $d_g=0$
(exact McNemar $p=2.16\times10^{-7}$).
The paired risk reduction is $8.67$ percentage
points (95\% seed-bootstrap CI $[5.33,12.00]$),
at mean paired time and path costs of
$+0.68$~s and $+0.290$~m.
Mean minimum clearance increases by $0.250$~m;
contact episodes are $7$ versus $35$ and
planar-overlap-proxy runs are $6$ versus $34$
(Markov versus baseline).
Seven baseline runs lack an evaluable freezing
metric. Shadow-window validity checks pass for
$299/300$ Markov and $291/300$ baseline runs,
without excluding contact-valid runs.
On $291$ common valid seeds, residual exceedances
are $1.006\%$ versus $52.591\%$
(not collision rates).
In the $299$ Markov-valid shadow runs,
$10/728$ $(j,h)$ cells exceed the $5\%$
design level, reaching $299/299$ at $j=0,h=8$.
The physical-contact improvement applies to
the generic cube; no pointwise probabilistic
or pedestrian-proxemic guarantee follows.

\subsection{Computational Evaluation of Semantics-Aware Obstacle Aggregation}

Two ROS~2/Gazebo studies benchmark grouping
ON/OFF from the same saved controller states
with $4$, $8$, and $16$ obstacles:
stationary generic cylinders
(A, $\tau_{\mathrm g}=1$~m) and moving pedestrians
simulated with HuNavSim 2.0, a ROS~2 human-navigation
simulator~\cite{escudero2027hunavsim2}
(B, $\tau_{\mathrm g}=2$~m).
During ON navigation, states are saved and
replayed in both modes from independent copies,
with two solves per mode in alternating
ON--OFF--OFF--ON or OFF--ON--ON--OFF order.
Reconstructed ON QPs are numerically verified
against the live objectives, constraints,
and bounds.
This yields $1541$ valid paired snapshots
from $58$ runs in A and $2042$ from $57$ in B;
two A runs lack valid replay and three B runs
end in Gazebo startup crashes.
With $G$ groups, the QP has $84+23G$
constraints and $22+2G$ variables.

\begin{table}[!htbp]
\centering
\caption{Same-state QP benchmark. Constraint and time
pairs are ON/OFF; timings include construction and solution.}
\label{tab:grouping_computation}
\scriptsize
\setlength{\tabcolsep}{2.0pt}
\begin{tabular}{ccrrrr}
\hline
Case & $n$ & Pairs & Constraints & Mean [ms] & p99 [ms] \\
\hline
A & 4  & 524 & 105/143 & 12.35/12.95 & 56.0/67.0 \\
A & 8  & 539 & 127/199 & 25.85/28.71 & 106.5/136.1 \\
A & 16 & 478 & 162/290 & 54.57/71.38 & 255.7/358.9 \\
B & 4  & 418 & 122/158 & 9.18/9.99   & 32.3/33.1 \\
B & 8  & 637 & 136/217 & 11.35/13.11 & 40.6/51.1 \\
B & 16 & 987 & 160/294 & 17.61/20.69 & 66.0/100.6 \\
\hline
\end{tabular}
\end{table}

In Table~\ref{tab:grouping_computation},
p99 denotes the 99th percentile of computation
times. Mean and p99 are computed from per-snapshot
averages of the two solves per mode.
At $16$ obstacles, grouping reduces constraints
by $43.9\%$ (A) and $45.7\%$ (B),
mean MPC time by $23.5\%$ and $14.9\%$,
and p99 by $28.7\%$ and $34.5\%$,
respectively.
Savings arise mainly from geometric QP
construction rather than Gurobi solution;
clustering costs average $0.092$~ms (A)
and $0.052$~ms (B).

\subsection{Closed-Loop Pedestrian Semantics and Personal-Space Compliance}

A separate ROS~2/Gazebo semantic ablation
uses $60$ nominal crowd scenarios per mode
($20$ each with $4$, $8$, and $16$ HuNavSim 2.0 pedestrians), with grouping ON and
$\varepsilon_g=0.30$.
Pedestrian mode uses asymmetric personal spaces,
whereas Generic uses ordinary obstacle geometry.
Nominal scenario inputs are shared, but live
pedestrian trajectories are not identically replayed.

To assess both modes by one common criterion, the recorded robot and
pedestrian poses and velocities are synchronized in simulation time.
At each sampled instant $t_k$, let $\mathcal F_R(t_k)$ denote the oriented
planar projection of Husky's nominal collision base and four wheels,
and $\mathcal P_j(t_k)$ the velocity-dependent personal space of
pedestrian $j$. We set $I_k=1$ if the intersection
$\mathcal F_R(t_k)\cap\bigcup_j\mathcal P_j(t_k)$ is a non-empty set,
and $I_k=0$ otherwise. Our primary personal-space intrusion ratio is
\begin{equation}\label{eq:psir_footprint}
\mathrm{PSIR}=100\,
\frac{\sum_{r,k}\Delta t_{r,k}\,I_{r,k}}
{\sum_{r,k}\Delta t_{r,k}}\,\%,
\end{equation}
where $\Delta t_{r,k}$ is the observed simulation-time interval in run
$r$. Simultaneous overlap with multiple pedestrians counts once in
this temporal ratio. The same offline personal-space evaluator is
applied to both modes, including Generic.

\begin{table}[!htbp]
\centering
\caption{Offline footprint--personal-space intersection (
simulation time). Run counts are Pedestrian/Generic.}
\label{tab:semantic_psir}
\scriptsize
\setlength{\tabcolsep}{4pt}
\begin{tabular}{cccc}
\hline
Pedestrians & Valid runs & Pedestrian [\%] & Generic [\%] \\
\hline
$4$  & $20/20$ & $1.35$ & $11.95$ \\
$8$  & $20/19$ & $1.82$ & $19.78$ \\
$16$ & $20/20$ & $3.04$ & $33.56$ \\
\hline
All  & $60/59$ & $\mathbf{2.11}$ & $\mathbf{22.84}$ \\
\hline
\end{tabular}
\end{table}

At least one sampled personal-space overlap
occurs in $32/60$ Pedestrian and $59/59$ usable
Generic runs, totaling $110.36$ and $927.80$~s
of overlap, respectively.
One Generic Gazebo failure provides no usable
navigation data and is excluded from PSIR.
These are geometric proximity outcomes,
not physical contacts; the planar footprint
excludes mesh-only collision components.

\section{CONCLUSIONS}

We integrated online Koopman prediction,
Markov-based residual and cone-screening margins,
and semantics-aware grouping into MPC.
Across $300$ paired generic-obstacle runs,
tightening reduced physical-contact runs
from $30$ to $4$, with an average $0.68$ s
travel-time increase.
Shadow comparisons expose conservatism
and onset exceedances, not finite-sample
coverage guarantees.
Grouping reduced same-state QP size and latency.
With grouping ON, offline footprint--personal-space
overlap was $2.11\%$ (Pedestrian) versus
$22.84\%$ (Generic) across $60$ attempted
scenarios per mode.
These geometric outcomes are not physical-contact
measurements or isolated effects of individual
mechanisms.


\section*{Acknowledgements}
This work was supported by Leonardo S.p.A. under Grant LDO/GMS\&I /P/0004870/26.



\bibliographystyle{IEEEtran}
\bibliography{references}

\end{document}